\documentclass{article}
\usepackage{spconf,amsmath,graphicx}

\usepackage{epstopdf}
\usepackage{subcaption}
\usepackage{amsfonts}
\usepackage{amsmath}
\usepackage{booktabs}
\usepackage{amsthm}
\theoremstyle{plain}

\usepackage{algorithm}
\usepackage{algpseudocode}
\usepackage{xcolor}
\usepackage{hyperref}

\def\N{{\mathcal{N}}}

\def \ltwo {$\ell_2$}
\def \th {\text{th}}
\def \z {\mathbf{z}}
\def \comment [#1]{\textcolor{blue}{\textit{Comment:} #1}} % comments in blue

\title{Variational Encoder-based Reliable Classification}
\name{Chitresh Bhushan*, Zhaoyuan Yang*, Nurali Virani, Naresh Iyer\thanks{*Equal contribution. \newline This work has been supported by GE Humble AI Initiative.}}
\address{GE Research, 1 Research Circle, Niskayuna, NY 12309}
\begin{document}

\begin{table*}[h]
	This is pre-print version of the article published in ICIP 2020:
	\vskip 0.5cm
	C. Bhushan, Z. Yang, N. Virani and N. Iyer, "Variational Encoder-Based Reliable Classification," 2020 IEEE International Conference on Image Processing (ICIP), Abu Dhabi, United Arab Emirates, 2020, pp. 1941-1945, doi: \href{https://doi.org/10.1109/ICIP40778.2020.9190836}{10.1109/ICIP40778.2020.9190836}
	
	\hrulefill
	\vskip 0.5cm
	\begin{center}
	\textcopyright\ 2020 IEEE. Personal use of this material is permitted. Permission from IEEE must be obtained for all other uses, in any current or future media, including reprinting/republishing this material for advertising or promotional purposes, creating new collective works, for resale or redistribution to servers or lists, or reuse of any copyrighted component of this work in other works.
	\end{center}
	
\end{table*}

\clearpage

%\ninept
%
\maketitle

\begin{abstract}
Machine learning models provide statistically impressive results which might be individually unreliable. To provide reliability, we propose an Epistemic Classifier (EC) that can provide justification of its belief using support from the training dataset as well as quality of reconstruction. Our approach is based on modified variational auto-encoders that can identify a semantically meaningful low-dimensional space where perceptually similar instances are close in \ltwo-distance too. Our results demonstrate improved reliability of predictions and robust identification of samples with adversarial attacks as compared to baseline of softmax-based thresholding.
\end{abstract}

\begin{keywords}
Classification, Justified Belief, Reliability, Interpretability, Adversarial Attacks
\end{keywords}
\section{Introduction}
\label{sec:intro}
Individual prediction reliability is key in safety-critical applications of machine learning (ML) in healthcare, industrial controls, and autonomy. 
To provide this reliability, the notion of epistemic classifiers (EC) was recently introduced in~\cite{virani2019JTBv1}. EC is a classifier that can justify its belief using support/evidence from neighborhoods in multiple layers. EC additionally provides exemplar-based interpretability using those supporting instances. In this paper, we propose epistemic encoders, where we co-train a variational auto-encoders (VAE) and a classifier to construct a low-dimensional semantically-meaningful embedding. The neighborhood support from training instances is then computed in that embedding to overcome curse of dimensionality and enforce agreement in \ltwo-distance and semantic similarity. The VAE can also provide reconstruction score at inference time, so we also use that as an additional \textit{support} in the justification process.

MagNet~\cite{meng2017magnet} uses autoencoder reconstruction error to either reject or reform potentially adversarial examples before the example is provided to a classifier. Unlike MagNet, where the autoencoder is trained independent of the classifier, our approach performs a joint training. In~\cite{kos_2018_adversarial_example_generative_models}, a classifier was trained on the latent space of a VAE to generate adversarial attacks for generative models. Here, we use a similar architecture with co-training to defend the network. Since the support operator in ECs is non-differentiable, it makes it less vulnerable to, and computationally more expensive for, white-box attacks compared to MagNet.

We make the following contributions: (a) an approach to identify a semantically meaningful low-dimensional space for computing support using \ltwo-distance; (b) introduce reconstruction quality as additional justification mechanism to identify uncertainties that neighborhood support cannot resolve by itself.

\section{Method}
\label{sec:method}

\subsection{Epistemic Classifiers (EC)}
\label{ssec:EpistemicClassifiers}

\begin{figure}[tb]
	\centering
	\includegraphics[width=6.3cm]{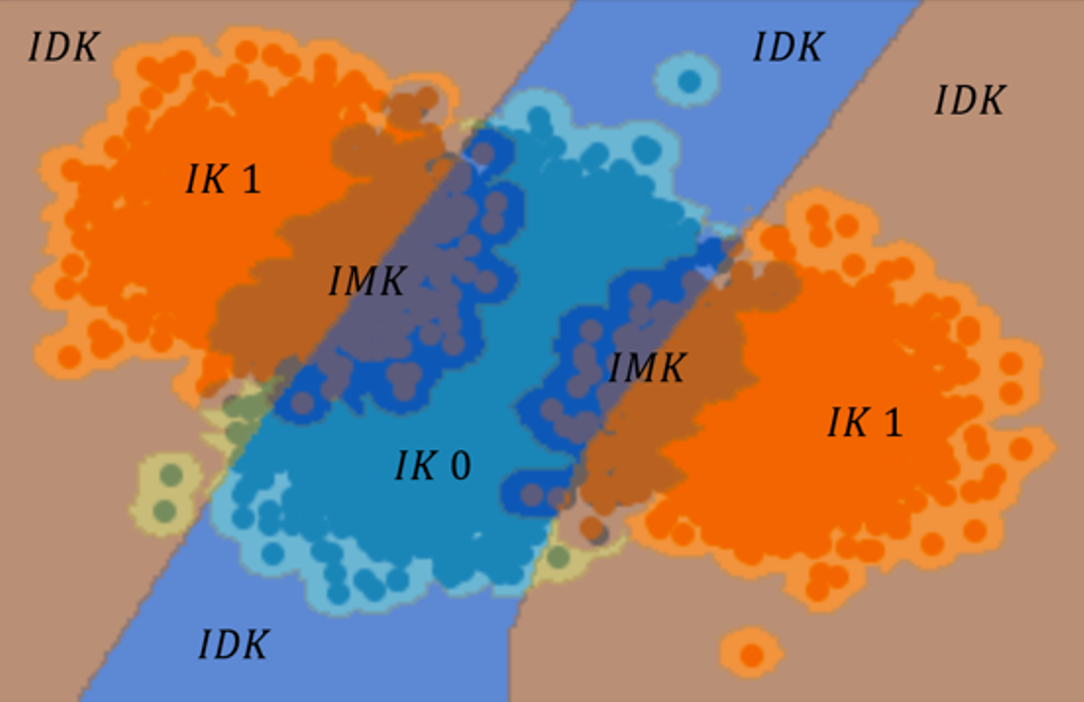}
	\caption{Illustration of region of: trust (IK0, IK1), confusion (IMK), and extrapolation (IDK) for 2D-input binary classification with Epistemic Classifier using a base NN classifier.}
	\label{fig:support}
\end{figure}

EC provides an approach to enhance prediction reliability for a classifier, which builds on the theory of justified true belief from epistemology~\cite{ichikawa2001analysis} and extends it to neural networks (NN)~\cite{virani2019JTBv1}. Specifically, ECs link reliability of predictions on a test input to characteristics of the support gathered from hidden layers of the network. For a given test sample $x$, ECs generate support for $x$ using training data as a mechanism to justify the class prediction for $x$. The support enables ECs to characterize the input space into: regions of extrapolation (“I don’t know” or IDK), regions of confusion (“I may know” or IMK), and regions of trust (“I know” or IK).
This enables annotating the classifier output (i.e. its belief) with IK, IMK, and IDK assertions (see Fig.~\ref{fig:support} for illustration). Traditional EC uses neighborhood-based support across multiple layers of a NN to obtain this justification.
The support $S_i(x)$ in $i^{\th}$ layer is defined as \cite{virani2019JTBv1}:
\begin{align}
S_i(x) &= \{f(\omega): \omega \in X, h_i(\omega) \in \N_i(h_i(x))\},
\label{eq:support}
\end{align}
where $f(\cdot)$ is function that maps training input to its training label, $h_i(\cdot)$ is the activation value in the $i^{\th}$ layer, and $\N_i(\cdot)$ is the neighborhood operator over the training data-set $X$.

The neighborhood operator $\N$ is generally defined using computationally tractable \ltwo-norm distance.
However, as most state-of-the-art classification networks use cascaded convolutional layers, a \ltwo-norm based distance metric for layer-activations does not necessarily reflect semantic or perceptual distance, especially in layers away from the output layer \cite{goodfellow_2016_DL_book, rawat2017CNN_classification_review}. Hence, use of \ltwo-norm based support from early layers can lead to an ill-informed justification of the belief (or output), causing the Epistemic classifier to assert IMK or IDK frequently in real-world application. In response, we propose ECs that use VAE to construct a semantically meaningful embedding for support generation and to augment support with reconstruction loss of the autoencoder. Next, we formally introduce this extension to ECs and show how it imparts better prediction reliability to ECs.

\subsection{Joint encoder-classifier approach}
\label{ssec:VAEclassifiers}

\begin{figure}[tb]
	\centering
	\includegraphics[width=7.5cm]{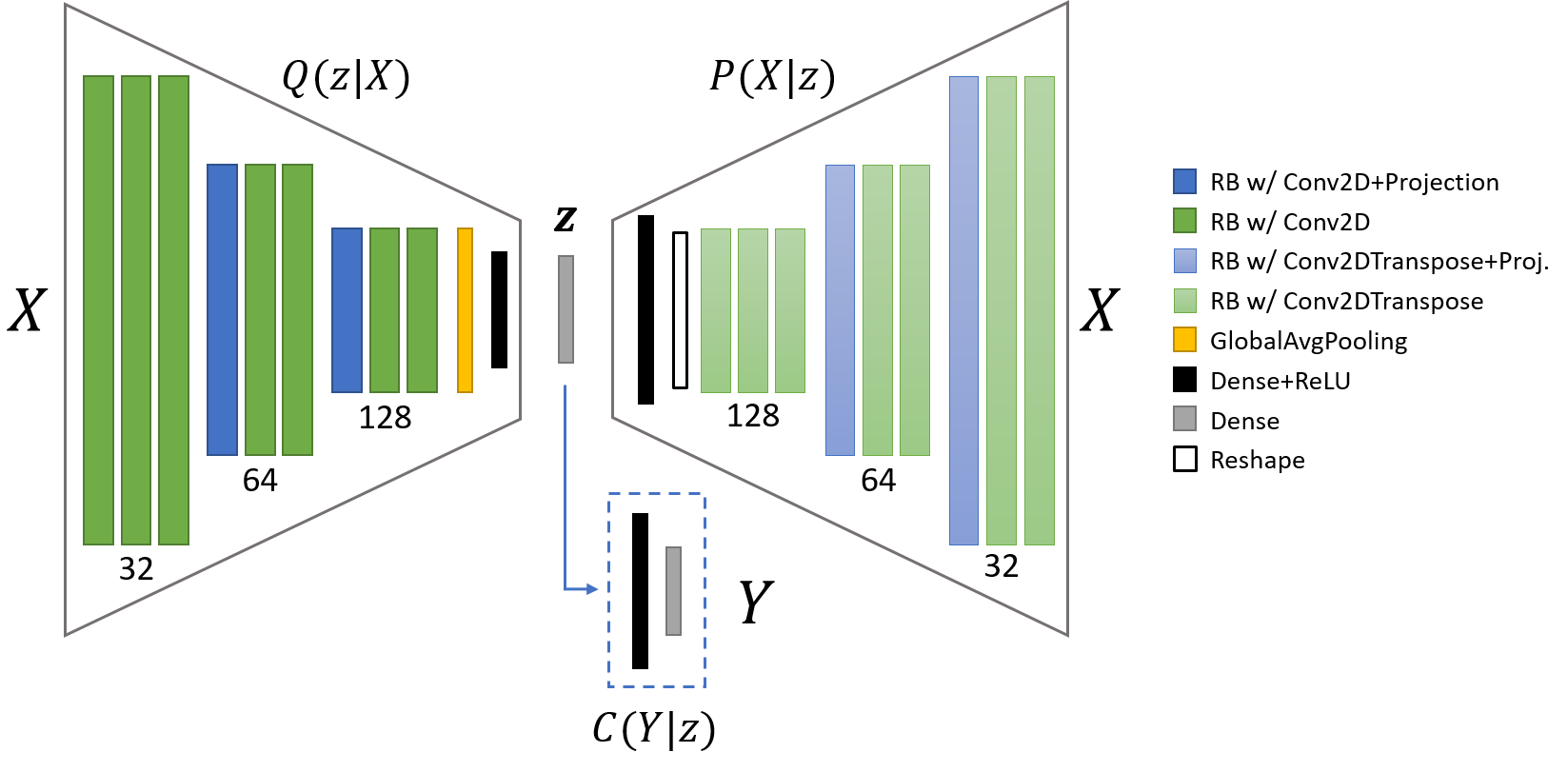}
	\caption{Outline of our modified VAE architecture that is co-trained with the classifier. `RB' stands for the standard residual block as described in \cite{he_2016_resnet}.}
	\label{fig:jtb2}
\end{figure}

Owing to VAE's excellent generative properties \cite{goodfellow_2016_DL_book,doersch_2016_VAE_tutorial,kingma_2013_VAE}, we hypothesize that the VAE's latent space is a perceptually meaningful space for support computation and retains ability to identify uncertainties without frequent IMK or IDK assertions. VAE are also known to produce unexpected reconstruction under adversarial attack \cite{kos_2018_adversarial_example_generative_models}, which further motivated use of VAE in this work.
Our EC share the `encoder' layers from VAE, as shown in Fig.\ref{fig:jtb2}, and is jointly trained with the modified VAE.

% why joint-VAE
VAEs use learned approximate Bayesian inference to generate a sample that is similar to the training set \cite{goodfellow_2016_DL_book,doersch_2016_VAE_tutorial,kingma_2013_VAE}.
It indirectly maximizes the model distribution $P(X)$ by minimizing the upper-bound of its negative log-likelihood \cite{kingma_2013_VAE,doersch_2016_VAE_tutorial,makhzani_2016_adversarial_VAE}.
In our approach we add another term to the upper-bound that captures the classification decoding capacity of learned  model distribution.
As shown in Fig.\ref{fig:jtb2}, lets assume $\z$ is the latent code vector of VAE, $Q(\z|X)$ is the encoding distribution, $P(X|\z)$ is the decoding distribution, $P(\z)$ is the prior normal distribution imposed in VAE, then we minimize following modified loss function for training the model.
\begin{equation}
\label{eq:VAE-joint-cost}
\begin{aligned}
	\mathcal{L} =             &  E_X[E_{Q(\z|X)}[-\log P(X|\z)]]                    &  & \text{(Reconstruction)}              \\
	              & + E_X[\text{H}(Q(\z|X),P(z))]                      &  & \text{(CrossEntropy($Q,P$))}         \\
	              & - E_X[\text{H}(Q(\z|X))]                           &  & \text{(Entropy($Q$))}                \\
	              & + \lambda_C  E_X[\text{H}(C(Y|\z), C^{\star}(Y|X))] &  & \text{(Classification)}
\end{aligned}
\end{equation}
where $C(Y|\z)$ is classification decoding distribution, $C^{\star}(Y|X)$ is the true classification decoding (known for training samples) and $\lambda_C$ is a scalar weight for the classification loss.
The first three terms are from~\cite{kingma_2013_VAE}, where first term can be interpreted as reconstruction loss of the VAE.
Rest of the terms can be seen as regularization terms for model optimization~\cite{makhzani_2016_adversarial_VAE}.
Second term (CrossEntropy($Q,P$)) encourages the posterior and prior to approach each other, which are chosen to be multivariate Gaussian in VAE.
Third term encourages the posterior to have non-zero variance, which should be helpful in avoiding badly scaled gradients during back propagation. 
The fourth term captures the categorical cross-entropy for classification and can be seen as another regularization term that pushes the VAE model to parameterize latent codes to be semantically-meaningful and suitable for classification.

\subsection{Justification: Support and Reconstruction}
\label{ssec:justification}

We consider both the quality of reconstructed outputs as well as the support $S$ for justification in our EC.
Unlike~\cite{virani2019JTBv1}, we use only the latent code of our VAE to compute the support for input $x$.
Specifically, we use eq.\eqref{eq:support} to compute support $S_\z(x)$, where $i$ has only one value corresponding to the encoder output with the latent code $\z$. Our neighborhood operator $\N(\cdot)$ is defined using $k$-NN neighborhood operation that identifies $k$ nearest (\ltwo-norm) training samples for an input in the latent space.

We use two different image dissimilarity metrics to estimate the loss $\mathcal{R}$ in the reconstructed output as compared to the input: mean-square error (MSE) and Structural Similarity Index (SSIM) \cite{wang_2004_SSIM}.
Function $\Phi(x, \tilde{x}, \bar{t})$ is used to identify the quality of reconstruction $\tilde{x}$, using thresholds $\bar{t}=\{t_{\text{MSE}}, t_{\text{SSIM}} \}$.
Reconstruction quality is identified as `Good' by function $\Phi$ if both of the losses are lower than corresponding thresholds, otherwise it is considered `Bad'.
Next we construct justification operator $J(x)$ as 
\begin{align}
J(x) = 
\begin{cases}
S_\z(x) & \text{if}\ \Phi(x, \tilde{x}, \bar{t}) = \text{Good}, \\
S_\z(x) \cup \phi & \text{otherwise},
\end{cases} 
\label{eq:justification} 
\end{align}
where $\phi$ is an arbitrary element to reflect bad reconstruction quality.
In other words, when the reconstruction error is low, $J(x)$ mirrors $S_\z(x)$ and the Justification set remains unchanged. However, when the reconstruction loss is high, uncertainty in support is increased by adding an arbitrary element.

\subsection{Algorithm and Implementation}
\label{ssec:algo}

\begin{algorithm}[tb]
	\caption{\textbf{--  Training VAE-based Epistemic Classifier}}
	\label{alg:EClassifier}
	\begin{algorithmic}[1] 
		\Require training set $(X, Y)$, validation set $(X^v, Y^v)$
		\Require trained modified VAE network $g$
		\Require distance metrics for latent code $d_\z$
		\State $\z^X \leftarrow$ Extract latent code $\z$ for training set $X$ 
		\State $\Omega \leftarrow \text{NeighborSearchTree}(\z^X, Y, d_\z)$
		%\State // Determine set of the parameters for justification
		\State $k, \bar{t} = \text{JustificationParameters}(X^v, Y^v, \Omega, g)$
		\State \Return $g, k, \bar{t}, \Omega$ 
	\end{algorithmic}
\end{algorithm}

\begin{algorithm}[tb]
	\caption{\textbf{--  Inference with our Epistemic Classifier}}
	\label{alg:EClassifier_Inference}
	\begin{algorithmic}[1]
		\Require Test input $x$
		\Require Epistemic classifier $G = (g,\mathcal{N}, \bar{t})$
		\State $(y^x, \tilde{x}) \leftarrow g(x)$  \Comment{Class \& reconstruction predictions}
		% \State $y^x \leftarrow$ Compute belief using class prediction $y$
		\State $\z^x \leftarrow$ Extract latent code $\z$ for input $x$
		\State $S_\z^x \leftarrow \Psi(\z^x, \mathcal{N})$ \Comment{support of $x$ in latent code}
		\State Get justification $J(x)$ using $S_\z^x, \Phi(x, \tilde{x}, \bar{t})$ and Eq.~\eqref{eq:justification}
		\If {$y^x = J(x)$}
		\State output $\leftarrow (\text{IK}, y^x)$
		\ElsIf {$y^x \subset J(x)$} \Comment{proper subset}
		\State output $\leftarrow (\text{IMK}, y^x)$
		\Else     \Comment{implies $y^x \not\subset J(x)$}
		\State output $\leftarrow (\text{IDK}, y^x)$	
		\EndIf
		\State \Return output
	\end{algorithmic}
\end{algorithm}

Our EC is build using a trained modified-VAE network $g$.
As shown in Fig.~\ref{fig:jtb2}, for an input $x$ network output has two elements, 
i.e. $g(x)=\{y, \tilde{x}\}$ where $y$ is label output and $\tilde{x}$ is the reconstructed output from latent code $\z$.
After the network $g$ is trained using the loss functions described in Sec.~\ref{ssec:VAEclassifiers}, the evidence for justification is derived from the training set $(X,Y)$ itself, as described in Algorithm.~\ref{alg:EClassifier}.
We extract latent codes across training set and construct a ball-tree $\Omega$ using a defined distance metric, which is represented by NeighborSearchTree function.
Ball-tree is used for nearest neighbor search \cite{liu_2006_highdim_k_nearest_neighbor} and it uses \ltwo-metric for computing distances.
JustificationParameters is a function that selects parameters for support operators. It selects the value of $k$ to define the neighborhood $\mathcal{N}$ for the support in latent space. In addition, it also computes a set of thresholds $\bar{t}$ using using $N^\th$ percentile of each metric across the validation set. Validation set is used to select value for $k$ and $N$. $\lambda_C$ was empirically set to a large value of 50 for all experiments.

During the inference stage, a testing sample $x$ is used with proposed EC according to Algorithm~\ref{alg:EClassifier_Inference}.
The belief of the classifier is same as the classifier output $y$.
The support of the input $x$ is computed by function $\Psi$, which uses its latent code and $\Omega$ to find $k$ nearest neighbors in training set. Function $\Phi$ quantifies the reconstruction quality using SSIM and MSE as described above.
Justification of the belief is then computed as per eq.~\eqref{eq:justification} using support and reconstruction quality.
This justification and belief is used to obtain the justified belief of our classifier. 
Note that justification set $J(\cdot)$ can be used to provide interpretable exemplars as evidence for the belief.

% details of architecture
We use residual blocks \cite{szegedy_2017_residual_connection_impact, he_2016_resnet} in encoder and decoder part of our modified VAE shown in Fig.~\ref{fig:jtb2}.
The classifier output is obtained by adding a single dense layer connection with ReLU activation to the latent vector $\z$ obtained from VAE. For classification part of the network, we use only the mean part of the latent code, which is a 16-length vector in our modified VAE.

\section{Experiments and Results}
\label{sec:results}

\begin{figure}[tb]
	\includegraphics[width=8.5cm]{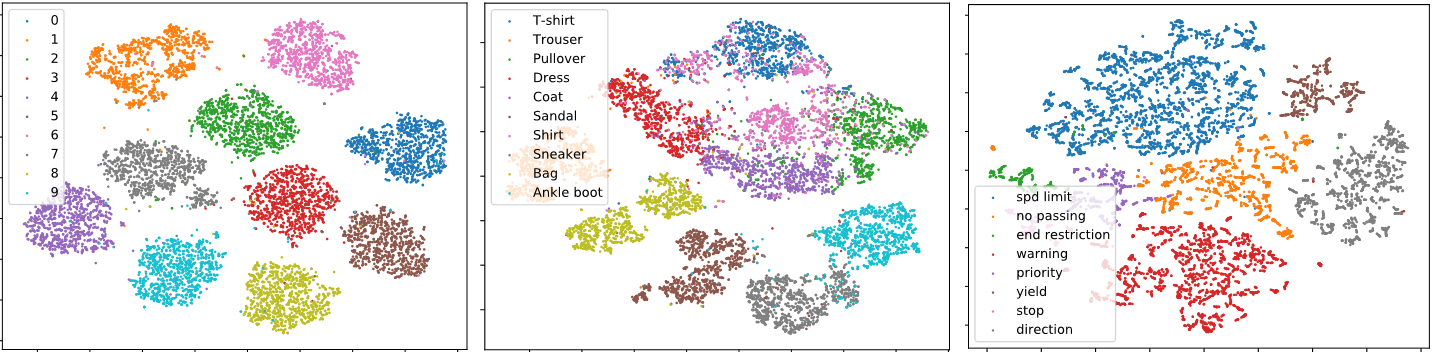}
	\caption{t-SNE visualization of the latent code $\z$ (length-16) for 
		(left) MNIST, (middle) Fashion-MNIST and (right) GTSRB dataset.}
	\label{fig:tsne}
\end{figure}

\begin{figure}[tb]
	\includegraphics[width=8.5cm]{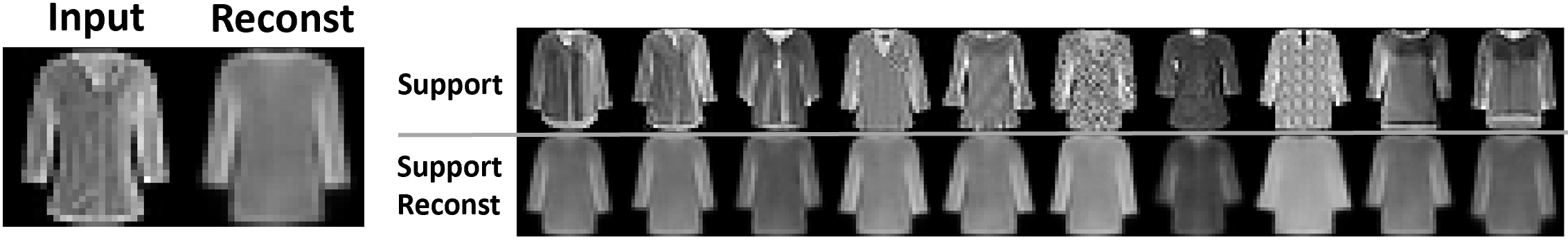}
	\caption{Example of reconstruction along with input's computed support from training set. Support's reconstruction is also shown in bottom row.}
	\label{fig:neighborhood-support}
\end{figure}

\begin{figure}[tb]
	\centering
	\includegraphics[width=8.5cm]{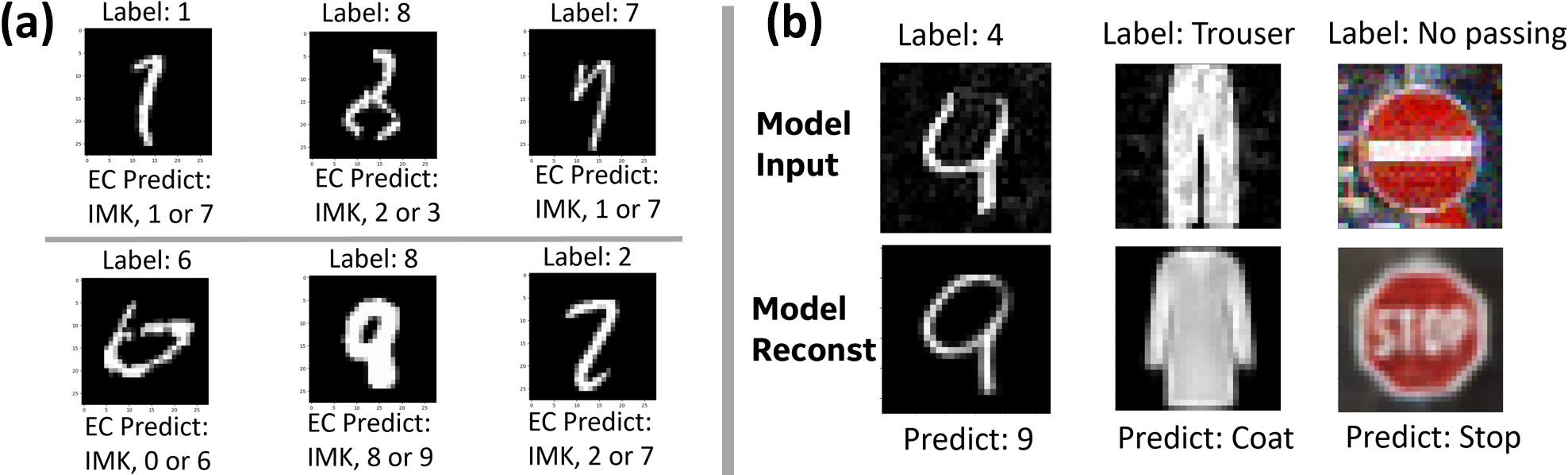}
	\caption{Examples of (a) confusion (IMK) along with their EC predictions, and (b) inputs affected by adversarial attacks along with corresponding reconstructed images.}
	\label{fig:attack-examples}
\end{figure}

\begin{table*}[tb]
	\centering
	\includegraphics[width=0.99\textwidth]{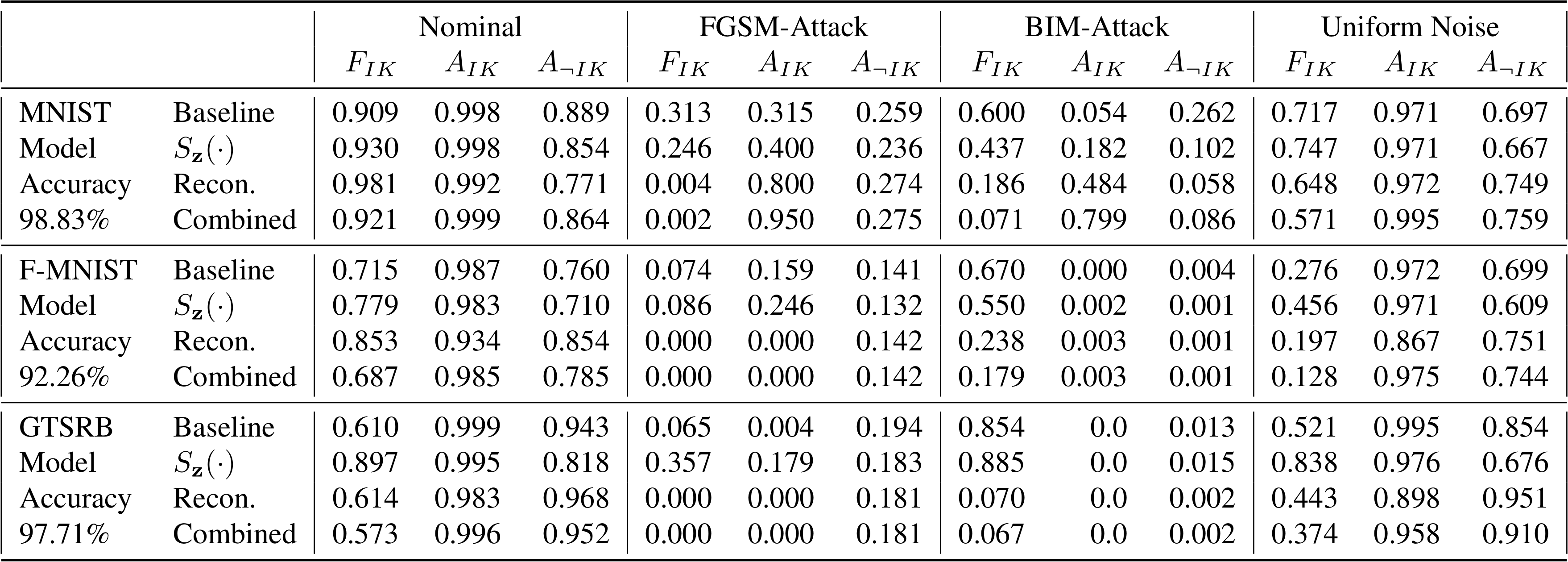}
	\caption{Performance of Epistemic Classifier on different dataset with different perturbation. Baseline is the performance using softmax thresholding. Base model test accuracy on nominal data is provided in the first column.}
	\label{table:result_all}
\end{table*}

We demonstrate usefulness of our approach with several data-sets (MNIST~\cite{lecun1998mnist}, Fashion-MNIST~\cite{xiao_2017_fashion_mnist} and German Traffic Sign Recognition Benchmark (GTSRB)~\cite{stallkamp2012man}) and test prediction reliability under various perturbations and adversarial attacks \cite{kurakin2016adversarial}.
Fig.~\ref{fig:tsne} shows t-SNE~\cite{maaten_2008_tsne_viz} visualization of the latent code $\z$ learned by our modified VAE for different datasets. 
It shows good separability across classes which is meaningful when computing support using \ltwo-norm distance metric and provides an effective computing space irrespective of dimensionality of inputs.
Fig.~\ref{fig:neighborhood-support} shows example of reconstruction and support for an input, demonstrating quality of the computed support.
Fig.~\ref{fig:attack-examples} shows a few IMK examples from MNIST test set and reconstruction for some inputs that were perturbed by BIM attack~\cite{kurakin2016adversarial}.
In adversarial attack cases, the reconstruction loss is high, which allows our EC to detect the attack.

Similar to~\cite{virani2019JTBv1}, we use augmented confusion matrix (ACM) to quantify performance of EC when testing with a dataset.
ACM consists of three sub-matrices, where each sub-matrix is a confusion matrix for predicted label versus true label under assertion of IK (top), IMK (middle), and IDK (bottom). 
In this work, we will use following three metrics derived from ACM to quantify performance: Coverage or fraction of IK ($F_{IK}$), accuracy over IK samples ($A_{IK}$), and accuracy over non-IK samples ($A_{\lnot IK}$). 
% clarify how F_IK, A_IK should be jointly interpreted.
In presence of an adversarial attack, a value of low $F_{IK}$ indicates that large fraction of cases were detected as attacked sample.
At the same time, a high $F_{IK}$ along with low $A_{IK}$ indicates poor performance in detecting the attack.
Further, a situation with $A_{\lnot IK} > A_{IK}$ indicates that the approach was too aggressive in classifying samples as non-IK or attack.
We use values of ($k$, $N$\%) as follows MNIST: (10, 99\%),  Fashion-MNIST:(50, 90\%), GTSRB:(50, 70\%).
Our justification approach is also compared to a baseline technique that uses thresholding on softmax outputs of the classifier.
For fair comparison, baseline thresholds were chosen to match $F_{IK}$ to our combined approach.
For adversarial attacks, we use FGSM \cite{goodfellow2014explaining} and BIM \cite{kurakin2016adversarial} with attack magnitude of 0.2 on the classifier outputs.
We also study effect of perturbation of inputs using uniform noise in range $[-0.1, 0.1]$.
For training, MNIST and Fashion-MNIST image were scaled to the range of $[0, 1]$.
Similar to~\cite{virani2019JTBv1}, we group GTSRB traffic sign dataset into eight types of traffic signs: speed limit, no passing, end of restriction, warning, priority, yield, stop and direction; yielding 34799 training, 4410 validation and 12630 testing images. 

\begin{figure}[tb]
	\centering
	\includegraphics[width=8.45cm]{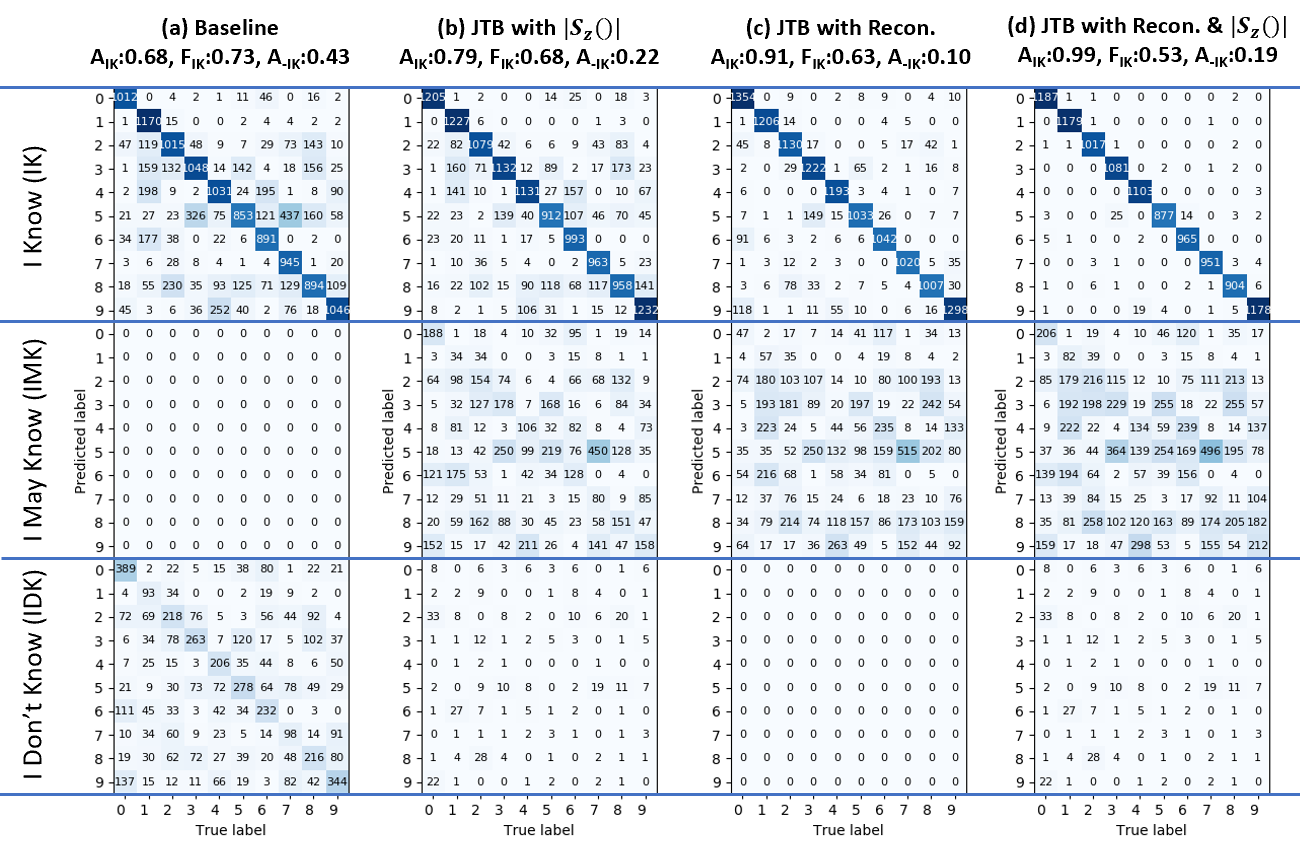}
	\caption{Augmented confusion matrices (ACM) for \emph{expanded} MNIST test-set using (a) Baseline, (b) only support-based, (c) only reconstruction-based, and (d) combined support \& reconstruction based justification. Half of the samples of \emph{expanded} MNIST test-set were perturbed by BIM-attack of 0.2 magnitude (see text for more details).}
	\label{fig:ACM}
\end{figure}

Fig.~\ref{fig:ACM} show ACMs with different forms of justification for an \emph{expanded} MNIST test set.
The \emph{expanded} MNIST test set was created by first perturbing each sample from the original MNIST test set with adversarial BIM-attack of 0.2 magnitude \cite{kurakin2016adversarial} and then appending these BIM perturbed samples to the original MNIST test.
This resulted into \emph{expanded} MNIST test set with twice as many samples as original MNIST test set, where one half of the samples were BIM-attacked.
A reliable classifier should achieve high accuracy for IK samples ($A_{IK}$) along with high coverage ($F_{IK}$).
Baseline justification approach using softmax thresholding (fig.~\ref{fig:ACM}a) achieves highest coverage of 73\%, however, with low IK-accuracy of 0.68. 
It also shows relatively high accuracy over non-IK samples ($A_{\lnot IK}$=0.43), which indicates that the used justification approach is sub-optimal, possibly resulting into several false negatives while identifying IK samples.
We achieve slightly better IK-accuracy of 0.79, when only support is used for justification (fig.~\ref{fig:ACM}b), however it is unable to identify several attacks.
Reconstruction based justification (fig.~\ref{fig:ACM}c) achieves higher IK-accuracy, however we achieve highest IK-accuracy of 0.99 when both support and reconstruction is used for justification (fig.~\ref{fig:ACM}d).
Our combined approach also achieves a coverage of 53\%, which is reasonable given that half of the samples of \emph{expanded} MNIST test set were BIM-attacked samples.
Note that when only reconstruction is used for justification, we define support in eq.~\ref{eq:justification} as $S_\z(x)=\{y^x\}$, where $y^x$ is the class prediction for input $x$. 
This implies that we cannot assert any belief as IDK when using reconstruction-only justification, as seen in fig.~\ref{fig:ACM}c.

Table~\ref{table:result_all} compares performance using all datasets with and without adversarial attack.
Baseline achieves good $F_{IK}$ and $A_{IK}$ for nominal testset but performs poorly with attacks (high $F_{IK}$ with low $A_{IK}$), which indicates that several attack samples were incorrectly classified with high confidence.
Our combined approach achieves high $F_{IK}$ with good $A_{IK}$ for nominal testset and shows excellent identification of attacks (low $F_{IK}$) by asserting samples as IMK/IDK (almost 100\% for FGSM and highest for BIM) across all data-sets.
Results for BIM-attack shows that $F_{IK}$ for combined approach is better in detecting attacks than either one of $S_\z$ or reconstruction by themselves.
This suggests that  $S_\z$ and reconstruction provide complementary information, as hypothesized. 
When inputs are perturbed by uniform noise baseline performs well, however, support-based approach achieves higher $F_{IK}$  with similar accuracy $A_{IK}$.

\section{Discussion and Conclusion}
\label{sec:conclusion}
In practice un-trustworthy (IDK/IMK) assertion from EC can be used to seek help from an expert and these parameters can be tuned to match the desired frequency to seek expert's help. Further, assertion of un-trustworthy classification into IMK and IDK can reduce the expert's effort to identify challenging cases. The computed support also provides a mechanism for obtaining training examples that the classifier believes is similar to test sample, which can be used for interpretability purposes.

Our current framework does not enforce any posterior distribution for each class in the latent space, which can change the ideal choice of support size for each class.
We plan to use approach described in \cite{makhzani_2016_adversarial_VAE} to enforce similar distribution for each class for more uniform effect of neighborhood size.
Our reconstructed images are smooth in nature, similar to other VAE~\cite{goodfellow_2016_DL_book}, which results in large SSIM loss and explains the low $F_{IK}$ in presence of large noise in table~\ref{table:result_all}.
In future, we will explore other dissimilarity metrics to address this issue.
Presented results show somewhat robust performance to gray-box or semi-white-box attack on the classifier. 
In future work, white-box attack on both classification and reconstruction will be studied.

In conclusion, we propose an Epistemic Classifier (EC) that can assert its belief based on justification from training set and shows robust performance to adversarial attacks. Our EC obtains semantically-meaningful latent space using modified VAE for support generation and uses reconstruction as an additional justification mechanism.

\bibliographystyle{IEEEbib}
\bibliography{refs}

\end{document}